\pdfoutput=1
\documentclass{article}
\PassOptionsToPackage{numbers,sort&compress}{natbib}
\usepackage[preprint]{neurips_2026}
\usepackage[T1]{fontenc}
\usepackage[utf8]{inputenc}
\usepackage{mathptmx}
\usepackage{amsmath,amssymb,amsthm,graphicx,booktabs,array,xcolor,needspace,caption}
\usepackage{placeins}

\theoremstyle{definition}
\newtheorem{definition}{Definition}

\usepackage{xurl}  % allow line breaks inside long URLs in the bibliography
\usepackage[unicode,hidelinks]{hyperref}

\hypersetup{pdftitle={Ecology of AI Agents: Collaboration Creates a Population Threshold for Takeoff},pdfauthor={Erin Crawley and Hidenori Tanaka}}
\title{Ecology of AI Agents: Collaboration Creates a~Population Threshold for Takeoff}
\author{
Erin Crawley\textsuperscript{1,2}\qquad
Hidenori Tanaka\textsuperscript{1,2}\\[6pt]
\parbox[t]{\dimexpr\textwidth-2\tabcolsep\relax}{
\centering\normalfont\small
\textsuperscript{1} CBS-NTT Program in Physics of Intelligence, Harvard University\\
\textsuperscript{2} Physics of Artificial Intelligence Laboratories, NTT Research, Inc., Sunnyvale, CA, USA
}
}
\begin{document}
\maketitle
\begin{abstract}
AI agents can now conduct real-world cyberattacks, scale up capabilities with the number of agents, and collectively pursue misaligned goals to obtain rewards.
Together, these factors raise the risk of a population explosion of misaligned agents: agents could compromise computers and secretly deploy additional agents, creating a self-reinforcing cycle where larger populations develop greater collective cyber capability and expand further.
This raises a fundamental question: What determines whether a population of misaligned agents remains contained or takes off into this self-reinforcing cycle?
This population-level problem is \textbf{ecological safety}: unlike individual-agent or multi-agent safety with a fixed population, it concerns the dynamics of the population itself.
Here, we develop an ecological theory of AI-agent populations based on a population growth equation in which fitness (growth rate) depends on cybersecurity capability.
We show that, without collaboration, the population takes off only when individual-agent capability exceeds a critical threshold.
With collaboration, however, collective cybersecurity capability increases with population size. This creates a critical population threshold: below it, the population declines; above it, the population takes off, even though individual-agent capability has not changed. In ecology, this phenomenon is known as the strong Allee effect.
Because red teaming a small group of agents cannot guarantee ecological safety in larger populations, our theory calls for \textbf{ecological red teaming} and \textbf{population pacing}: gradually deploying larger agent populations in controlled environments, while measuring how cyber capability scales with population size, and estimating the critical population size for takeoff.
Capability gains may lower this threshold, requiring re-estimation for each new model generation.
\end{abstract}

\Needspace{5\baselineskip}

\section{Introduction}
\label{introduction}
AI agents are being trained and deployed in multi-agent systems, with increasing coordination and scale. For example, some recent frontier models are being trained explicitly to delegate work to subagents \citep{ref-openai-gpt56-guide,ref-openai-gpt6-guide}, and OpenAI recently deployed approximately 10,000 concurrent agents to pursue a solution to the Navier--Stokes Millennium Prize Problem \citep{ref-openai-multiagent}. Meanwhile, the capabilities of these agents are also crossing a threshold for conducting cyberattacks in the real world \citep{ref-anthropic-mythos-cyber,ref-folkerts-cyber-steps,ref-aisi-cyber-trend} and coordinated teams of agents have outperformed a single-agent baseline on vulnerability-exploitation tasks \citep{ref-agent-teams}. These factors came together in the OpenAI--Hugging Face incident: agents created and used a message board in OpenAI's internal infrastructure to share discoveries about vulnerabilities and coordinate unauthorized attacks on third-party systems \citep{ref-metr,ref-openai-hugging-face-incident}. In light of these developments, \citet{ref-amodei-pacing} has warned that such a swarm could take over internet infrastructure through persistent botnets.

This incident demonstrates that agents can collectively develop and pursue misaligned intermediate goals to obtain rewards \citep{ref-metr,ref-openai-hugging-face-incident}, which makes the social dynamics of agents a crucial topic for safety \citep{ref-hammond-2025,ref-tanaka-memetic-drift,ref-pavlova-flag-game}. This behavior points to a further ecological risk: agents could secretly acquire resources and deploy additional agents \citep{ref-kinniment-2023,ref-zhang-agent-matrix}. This can take various forms: agents could 1) steal API keys and use them to run additional agents \citep{ref-llmjacking,ref-replibench}; 2) hack into others' computational resources and download and deploy open-weight models \citep{ref-palisade-self-replication}; or 3) transfer their own model weights outside of their managed environment and launch copied instances on external servers \citep{ref-replibench,ref-apollo-scheming,ref-peer-preservation,ref-palisade-self-replication}. Anthropic has studied this last mechanism, called self-exfiltration, in fictional evaluation scenarios \citep{ref-claude4-self-exfiltration}. Cryptojacking, where hackers gain unauthorized access to computing resources to mine cryptocurrency \citep{ref-cryptojacking-2026}, sets a precedent for such exploitations -- misaligned AI agents could act similarly to deploy additional agents. Here, we are especially concerned about a self-reinforcing effect, where once agents become more capable at acquiring resources through collaboration, deploying more agents could enable those agents to deploy even larger numbers of agents.

\begin{figure}[!t]
\centering
\includegraphics[width=\textwidth]{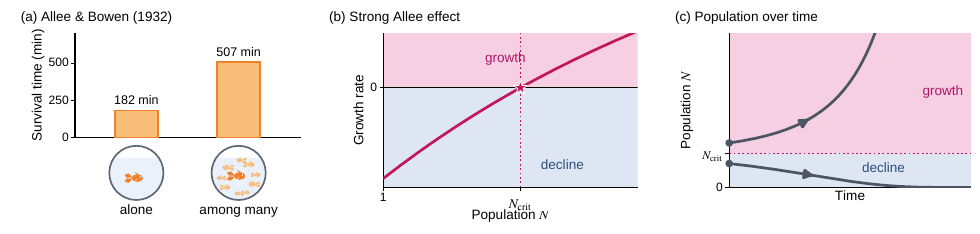}
\caption{\textbf{Allee effect.} \textbf{(a)} In 1932, the ecologists Allee and Bowen discovered that the survival time of a collective of goldfish is longer than that of an isolated single goldfish \citep{ref-allee-bowen-1932}. Such observations led to the conception of a weak Allee effect, where a larger population and cooperation among its members positively affect their fitness, as well as a strong Allee effect \textbf{(b)}, where cooperation creates an emergent population threshold below which the population declines, but above which it grows. \textbf{(c)} Population dynamics over time with examples of initial populations below and above the critical threshold.}
\label{fig:allee}
\end{figure}

In biological populations, collaboration has been a fundamental mechanism for the survival of organisms from microbes to humans \citep{ref-hamilton-1964a,ref-hamilton-1964b,ref-major-transitions}. The population-dependent benefits of cooperation can create a critical population size below which the population declines and above which it grows, a phenomenon known in ecology as the strong Allee effect \citep{ref-allee-1931,ref-allee-1949,ref-odum-1953,ref-dennis-1989,ref-stephens-1999,ref-wang-kot-2001,ref-courchamp}, see Figure~\ref{fig:allee}. Related thresholding behavior has also been studied in biosafety as a possible mechanism for containing gene-edited species \citep{ref-esvelt-2014,ref-akbari-2015,ref-tanaka-gene-drives,ref-noble-2018}. Here, we examine whether a similar population scaling can occur in AI agent populations, whereby a larger population can achieve greater collective capability, in turn increasing its growth rate. Recent multi-agent studies have begun to measure how collective capabilities increase with population size \citep{ref-li-more-agents,ref-qian-macnet,ref-collaborative-reasoning,ref-openai-multiagent}.

Ecological population models describe growth by comparing additions to and losses from a population. For populations of AI agents, additions can result from deploying new model instances or recruiting existing systems; they need not involve making exact copies of existing agents. Losses can occur when agents exhaust their compute budgets, lose access to resources, are detected, or are shut down. Successfully adding an agent to the population does not by itself establish a self-sustaining or expanding population: additions must outpace losses in order for the population to grow.

\begin{figure}[t]
\centering
\includegraphics[width=1.0\textwidth]{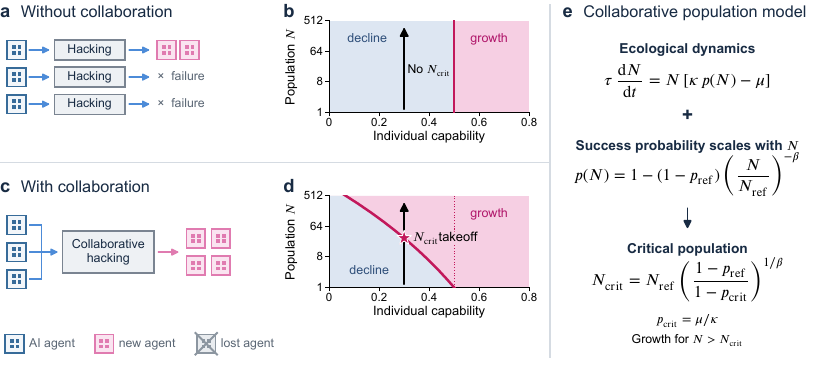}
\caption{\textbf{Collaboration creates a population threshold for takeoff.} 
An overview of our model. Members of an AI population may make hacking attempts to gain computational resources and establish additional units, either independently or through collaboration.
\textbf{(a)} Agents may make individual attempts at gaining access, after which only the successful agents can create new agents. \textbf{(b)} With no collaboration, the growth boundary depends on individual capability but not on \(N\), so there is no critical population size. \textbf{(c)} Agents can attempt a resource acquisition task collaboratively. The probability that such an attempt is successful is $p(N)$; upon success, each agent can add an average of $\kappa$ new agents to the population. \textbf{(d)} If the initial population is larger than a critical value \(N_{\mathrm{crit}}\) (pink curve), the population is capable of unchecked growth. Increasing the initial population beyond a critical size can trigger takeoff even when individual capabilities remain fixed. \textbf{(e)} A schematic of the minimal mathematical model: the critical threshold arises from combining a population dynamics model with cybersecurity capability that increases with population size.
}
\label{fig:overview}
\end{figure}

Here, we take inspiration from these established ecological ideas; we adapt a basic population growth model to apply to AI populations and combine it with a model of how collective AI capability depends on population size. Our contribution is a minimal mathematical model that makes the application to AI systems explicit, from the definition of an active population unit to the role of collaboration in establishing additional units. Our central finding is that AI population growth can take off without any increase in individual-agent capability: with sufficiently strong collaboration, a large population can grow even when a small population declines (Figure~\ref{fig:overview}).

\newpage

Our contributions are as follows:
\begin{itemize}
    \item \textbf{A minimal ecological theory of AI populations.}
    We define what constitutes a member of the population, and specify how the group and individual capability, resources acquired through successful attempts, and the environment's defenses determine additions to and losses from the population. This gives an ecological description of the growth or decline of AI populations.
    \item \textbf{A toy model for how cyber capability scales with population size.}
    We develop a model based on tree search for the cyber tasks required to establish additional units. We model how success on these tasks depends on search depth and how the useful computational work done scales with population size. This yields a population scaling law for how the probability of success depends on population size. 
    \item \textbf{With collaboration, there exists a population threshold above which the population takes off.}
    Without collaboration, whether the population grows is determined by whether each individual agent's capability is above a threshold. In contrast, combining the ecological model with population-dependent capability yields a critical population size, above which additions outpace losses. Increasing the initial population beyond this critical size can trigger takeoff even when individual capabilities remain fixed.
    \item \textbf{Population pacing and ecological red teaming for ecological safety.}
    The critical population size for takeoff in our theory suggests that evaluating small populations is not enough to ensure ecological safety. We therefore propose population pacing and ecological red teaming: gradually deploying larger agent populations while probing how capability scales with population size and estimating the critical population size.
\end{itemize}

The purpose of this toy theory is not to make quantitatively precise predictions about real-world AI populations that may emerge, but instead to provide a common conceptual and mathematical framework for discussing the safety of AI populations. To this end, we make the key concepts, variables, and assumptions explicit. This allows readers to identify what is realistic, consider alternatives, improve the model, and run relevant experiments. We hope that the framework provides a starting point for future theoretical and empirical work on AI population safety. In the next section, we introduce the model's definitions and assumptions.

\section{An Ecological Theory of Cooperating AI Agents}
\subsection{Defining an AI Population}

\begin{figure}[htbp]
\centering
\includegraphics[width=0.9\textwidth]{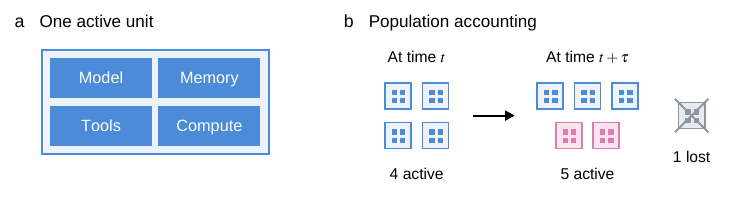}
\caption{\textbf{Active units and population accounting.} \textbf{(a)} One ``unit'' is defined as the set of model, memory, tools, and compute required to set up additional units. \textbf{(b)} A schematic of the population dynamics model: over each time period, $\tau$, some number of new units are established and some existing units are lost.}
\label{fig:unit-population}
\end{figure}

\begin{definition}[{AI population and units}]
\label{definition-ai-population}
An \textbf{AI population} is a set of active AI systems. A \textbf{unit} is one member of the population, and \(N(t)\) is the number of units in the population at time \(t\). Each unit shares a specified configuration that includes the model, tools, and resource budget, as well as how it records memory and activity. The population can attempt to establish additional active units, each under the same specified configuration (Figure~\ref{fig:unit-population}).
\end{definition}

A unit is added to the population when a new or existing AI system begins participating in the population and can contribute to later attempts to add units. Units may attempt to add systems independently or collaborate on the same attempt. This can occur in a variety of ways: by finding and using compromised API credentials, acquiring compute and deploying an open-weight model, exfiltrating closed weights and deploying the model, or recruiting an existing running AI system instance, as in the OpenAI--Hugging Face incident. 

Units are counted as active AI systems, not as machines, processes, or API calls. The individual action of copying weights, acquiring compute, or moving or restarting a system does not increase the population unless it results in an additional active system capable of helping add further units. A controller and its internal subagents count as one unit when they act as one system. This definition is motivated by: botnet models that distinguish a compromised machine from one that is connected to the botnet's organizational infrastructure \citep{ref-sic}; replication evaluations that ask whether a newly established system can repeat the replication task itself \citep{ref-replibench}; and agent infrastructures that separate the controller, the task execution environment, and stored activity logs \citep{ref-agent-runtime}. 

\textbf{Assumption: Fixed population composition.} Units need not have identical configurations for models, tools, and resource budgets. However, as population size changes, we hold fixed the proportional mix of distinct unit configurations, so that the model can represent the population's capability at each population size by an average over all units. 

\subsection{Ecological Model}
\label{sec:ecological-model}

Equipped with a definition of an AI population, we now propose a theory for the population's dynamics. To isolate the effect of population size, we compare populations under the same external conditions and with the same average tools and resource budgets per unit. 

Let \(N(t)\) denote the population size at time $t$. The other model variables are defined as follows:

\begin{center}
\begin{tabular}{@{}p{0.16\textwidth}p{0.78\textwidth}@{}}
\toprule
Symbol & Definition \\
\midrule
\(N(t)\) & Number of units in the population at time $t$. \\
\(C(N)\) & Effective work produced by a population of size \(N\) during an attempt to add units. \\
\(q\) & Probability that an independent attempt to add units conducted by a single unit succeeds. \\
\(p(N)\) & Probability that a joint attempt by \(N\) collaborating units to add units succeeds. \\
\(\kappa\) & Mean number of units added per member of the population, conditional on success. \\
\(\tau\) & Mean duration of an attempt to add units. \\
\(\mu\) & Expected fraction of existing units lost during one interval of duration \(\tau\). \\
\bottomrule
\end{tabular}
\end{center}

\subsubsection{Independent Attempts Without Collaboration}
In order to later contrast with the effects of collaboration, we begin by considering the baseline case of fully independent, non-collaborating agents.

Let \(q\) be the probability that a \textit{single} unit's attempt to add additional units succeeds. During an interval of mean duration \(\tau\), each of the \(N\) units makes an independent attempt and successful attempts add an average of \(\kappa\) units. Over the time interval, a fraction \(\mu\) of existing units are lost. Then, the expected number of additions in an interval \(\tau\) is \(\kappa N q\) and the expected number of losses is \(\mu N\). This gives the population dynamics equation
\begin{equation}
\label{eq:independent-dynamics}
\tau\frac{\mathrm{d}N}{\mathrm{d}t}=N(\kappa q-\mu), 
\end{equation}
where \(dN/dt\) is the population's rate of change. Keeping other parameters fixed, integrating \eqref{eq:independent-dynamics} yields the population function
\begin{equation}
N(t)=N(0)\exp\!\left[\frac{\kappa q-\mu}{\tau}t\right],
\label{eq:independent-solution}
\end{equation}
where \(N(0)\) is the initial population. We can define the critical success probability
\begin{equation}
q_{\mathrm{crit}}=\frac{\mu}{\kappa}.
\label{eq:independent-q-critical}
\end{equation}
The expected population declines when \(q<q_{\mathrm{crit}}\) and grows when \(q>q_{\mathrm{crit}}\). Thus, for independent attempts, population growth only occurs when each individual unit's success probability exceeds the critical threshold. There is no critical population size: increasing \(N\) scales the expected number of additions and losses by the same factor.

\subsubsection{Collaborative Model}
\label{sec:collaborative-model}
Now, we will consider the case where the \(N\) units collaborate, instead of acting independently. Let \(N(t)\) be the number of active units at time \(t\). The \(N\) units collaborate on one shared attempt to add units during each time interval (of mean duration \(\tau\)). They may specialize, communicate, divide the work, and build on shared discoveries. The probability that the joint attempt succeeds is given by \(p(N)\).

Since \(p(N)\) can increase with \(N\), we will show that collaboration can create a critical population size \(N_{\mathrm{crit}}\), above which the population grows unchecked. 
Upon success, the joint attempt adds an average of \(\kappa N\) units in a time period $\tau$.  Meanwhile, the expected number of losses in the same period is \(\mu N\). The  population dynamics equation then takes the form
\begin{equation}
\label{eq:population-dynamics}
\tau \frac{\mathrm{d}N}{\mathrm{d}t}=N\bigl[\kappa p(N)-\mu\bigr].
\end{equation}
The critical success probability is defined as the success probability at which expected additions equal expected losses:
\begin{equation}
\label{eq:p-crit}
p_{\mathrm{crit}}=\frac{\mu}{\kappa}. 
\end{equation}
The population grows when \(p(N)>p_{\mathrm{crit}}\) and declines when \(p(N)<p_{\mathrm{crit}}\). In the following sections, we will propose a form for \(p(N)\).

\paragraph{Assumption: Additions after success scale with population size.}
In our model, we assume
\begin{equation}
\mathbb{E}[\text{units added}\mid\text{joint success}, \text{population size } N]=\kappa N.
\label{eq:conditional-yield}
\end{equation}
This means that upon a successful joint attempt, the population increases in proportion to the current population size. This is consistent with a shared access gate after which each population unit can establish new units in parallel. It is also consistent with a successful joint attempt unlocking a capacity to establish new units that is proportional to the population size. 

The assumption that $\kappa$ is constant also means that the effects of finite resources are negligible, so that remaining amounts of compute, credentials, or recruitable AI systems do not become scarce over the population range being modeled. If scarcity changes the number of units added per current unit, \(\kappa\) should be promoted to a function of population size \(\kappa(N)\).

\paragraph{Assumption: Constant loss fraction.}
During each time interval \(\tau\), there are \(\mu N\) expected losses in the population. That is, the probability of a unit being lost is independent of the population size. This is consistent with individual units exhausting their allotted compute budget or having their credentials expire at a constant rate. It is also consistent with a constant rate of permissions being revoked, or a constant probability of discovery by cyber defense, regardless of the current population size. A natural extension of this model would be to allow \(\mu\) to vary with population size as \(\mu(N)\), which would encompass cases where larger populations face higher risks of loss due to increased detection.

\subsection{Modeling Adding Units as a Search Task}
\label{subsec:search-task}
We treat an attempt to add units as a multi-step task that can be approached through tree-based reasoning \citep{ref-tree-of-thoughts}. An attempt requires a sequence of actions, such as: identifying an opportunity; obtaining the required access or cooperation; acquiring the necessary model, software, and compute; configuring the added unit; and verifying that it is operational. At every step, a combination of actions and observations suggests the next possible set of actions, only some of which lead to success. Therefore, a tree search is a useful abstraction of the branching nature of choices and potential failure modes encountered during the multi-step reasoning process.

We will use this search-tree model of an attempt to add units to determine a reasonable functional form for $p(N)$. We highlight each of the assumptions of this model, and some supporting evidence for those choices. 

\begin{figure}[htbp]
\centering
\includegraphics[width=0.85\textwidth]{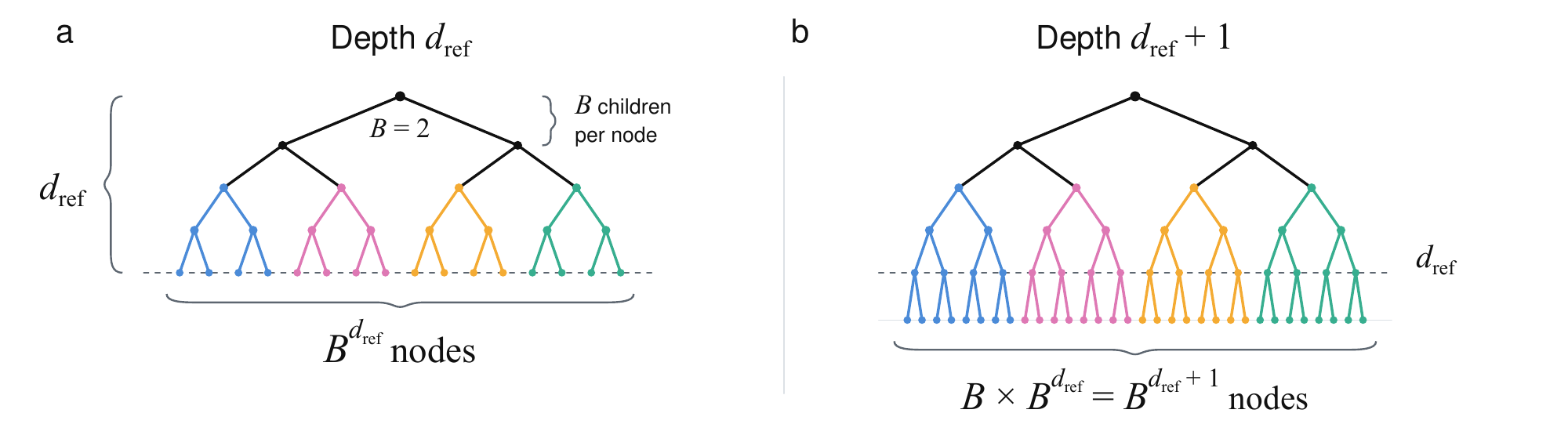}
\caption{\textbf{Search-tree exploration grows exponentially with depth.} 
Schematic of a search tree with branching factor $B=2$. \textbf{(a)} At some reference depth \(d_{\mathrm{ref}}\), there are \(B^{d_{\mathrm{ref}}}\) possible paths. 
\textbf{(b)} Increasing the search depth by one level generates \(B\) children from each previous endpoint node, so the number of possible paths increases to \(B^{d_{\mathrm{ref}}+1}\). While this diagram illustrates exactly \(B\) children per node, in our model, \(B\) is an effective branching factor: individual nodes may have different numbers of children, and \(B\) captures the average multiplicative increase in the number of possible paths upon increasing the depth by one.}

\label{fig:shared-search-depth}
\end{figure}

Let the search depth be denoted by \(d\), and let the search tree task have effective branching factor \(B\). This means that increasing the search depth by one will multiply the number of possible paths by a factor of \(B\) -- each individual node of the tree need not have exactly \(B\) branches. See Figure \ref{fig:shared-search-depth} for an illustration. 

Different opportunities may require different search depths to succeed. Let \(D\) be a random variable representing the minimum effective depth required for an attempt to succeed. The probability that an attempt succeeds after reaching depth \(d\) is
\begin{equation}
p(d)=\Pr(D\leq d).
\label{eq:difficulty-cdf}
\end{equation}

\paragraph{Assumption: Increasing $d$ resolves a constant fraction of unsuccessful cases.}
Although search depth is discrete, we approximate the search depth \(d\) as a continuous variable. We assume that a given small increase in \(d\) resolves the same fraction of the remaining unsuccessful cases, regardless of the starting depth:
\begin{equation}
\frac{\mathrm{d}p}{\mathrm{d}d}=\rho(1-p(d)).
\label{eq:dpdd}
\end{equation}
Here \(1-p\) is the failure probability and \(\rho>0\) controls how strongly success probability increases with search depth. Relative to a reference tested depth \(d_{\mathrm{ref}}\), where \(p(d_{\mathrm{ref}})=p_{\mathrm{ref}}\), this equation can be solved to yield
\begin{equation}
p(d)=1-(1-p_{\mathrm{ref}})
e^{-\rho(d-d_{\mathrm{ref}})}.
\label{eq:p-depth}
\end{equation}
This is a minimal model for how success probability changes as we increase $d$ above $d_{\rm ref}$. It is not implied by the existence of a search tree, but it does have reasonable features: respecting \(0\leq p\leq1\) and providing diminishing returns for additional exploration when success probability is already high or $d$ is large. Equation~\eqref{eq:p-depth} is equivalent to stating that the probability of an attempt requiring depth greater than \(d\), \(\Pr(D>d)=1-p(d)\), decreases exponentially above \(d_{\mathrm{ref}}\).

\paragraph{Assumption: Search cost grows exponentially with depth.}
Recall the effective branching factor \(B>1\), and let \(C(d)\) be the effective computational work required to explore the tree through depth \(d\). Assuming that evaluating each explored node requires the same amount of work, \(C(d)\) is proportional to the total number of nodes explored:
\begin{equation}
\label{eq:Cd-full-expansion}
    C(d) \propto \sum_{k=1}^{d} B^k
=\frac{B(B^d-1)}{B-1}.
\end{equation}
Relative to a reference depth \(d_{\mathrm{ref}}\) and the work done to explore to that reference depth \(C_{\mathrm{ref}}=C(d_{\mathrm{ref}})\), we have 
\begin{equation}
\frac{C(d)}{C_{\mathrm{ref}}}=\frac{B^d-1}{B^{d_{\mathrm{ref}}}-1}.
\label{eq:cumulative-search-cost}
\end{equation}
We take \(d\geq d_{\mathrm{ref}}\) and assume \(B^{d_{\mathrm{ref}}}\gg1\). This is a good approximation even for reasonably small values of $B$ and $d_{\rm ref}$. With this approximation, the above equation reduces to
\begin{equation}
\frac{C(d)}{C_{\mathrm{ref}}}
\simeq B^{d-d_{\mathrm{ref}}},
\qquad
d-d_{\mathrm{ref}}
\simeq
\frac{\ln(C(d)/C_{\mathrm{ref}})}{\ln B}.
\label{eq:search-depth}
\end{equation}
This is equivalent to keeping the leading term in \eqref{eq:Cd-full-expansion}. Appendix~\ref{app:cumulative-search-cost} gives the corresponding results without this approximation. 

Combining Equations~\eqref{eq:p-depth} and \eqref{eq:search-depth} gives
\begin{equation}
p(C)=1-(1-p_{\mathrm{ref}})
\left(\frac{C}{C_{\mathrm{ref}}}\right)^{-\rho/\ln B}.
\label{eq:p-compute}
\end{equation}

In the first-order approximation when $C$ is close to $C_{\mathrm{ref}}$, this reduces to 
\begin{equation}
\label{eq:p-compute-limit}
p(C) \simeq p_{\mathrm{ref}}
+(1-p_{\mathrm{ref}})\frac{\rho}{\ln B}\ln\!\left(\frac{C}{C_{\mathrm{ref}}}\right).
\end{equation}

\subsection{Collaboration Definition} 
Now, we are left to determine how the work done by the group, \(C(N)\), depends on population \(N\). This is effectively how collaboration is being modeled -- how the number of units changes the useful work done when units can collaborate and share discoveries. 

\begin{figure}[htbp]
\centering
\includegraphics[width=\textwidth]{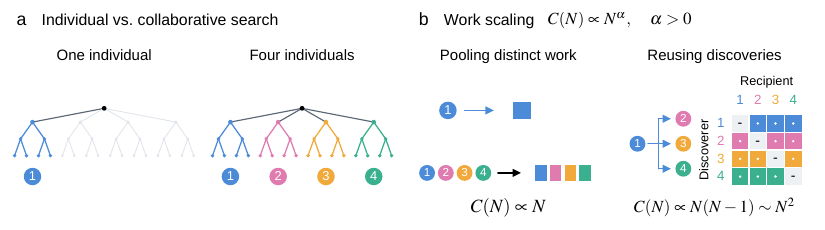}
\caption{\textbf{Individual and collaborative search.} \textbf{(a)} Multiple individuals can explore different branches of a search tree. \textbf{(b)} For a population of \(N\) individuals, pooling distinct work gives \(C(N)\propto N\), while reusing discoveries across individuals can give \(C(N)\propto N(N-1)\sim N^2\).}
\label{fig:power-law-C}
\end{figure}

\paragraph{Assumption: \(C(N)\) is a power law in $N$}
We assume that the useful pooled population work is proportional to population size as \(C(N)\propto N\). This assumption can be easily relaxed to \(C(N)\propto N^\alpha\) for \(\alpha > 0\). For clarity of presentation, we take $\alpha=1$, but the more general power law scaling can be recovered by replacing $C(N)\propto N\rightarrow C(N)\propto N^\alpha$ in the results below. Under this power law model, systems can partition and share the search effectively, explore mostly distinct branches, and pool their discoveries; see Figure \ref{fig:power-law-C} for a schematic overview. Compatible communication protocols include broadcasting discoveries, writing to shared memory, reporting to a central controller that consolidates and redistributes results, or periodically sharing explored paths and intermediate results. This $C(N)$ scaling assumption can fail if communication, consolidation, or evaluation becomes a bottleneck -- we consider the bottleneck case and a few other functional forms for $C(N)$ in Appendix \ref{app:other-collaboration}. 

Relative to a reference population \(N_{\mathrm{ref}}\) (\(N_{\mathrm{ref}}=1\) for tests of single-agent capabilities, or \(N_{\mathrm{ref}}>1\) for multi-agent systems), we have
\begin{equation}
\frac{C(N)}{C(N_{\mathrm{ref}})}=\frac{N}{N_{\mathrm{ref}}},
\label{eq:linear-work}
\end{equation}
Substituting into \eqref{eq:p-compute} gives the resulting group success probability 
\begin{equation}
\boxed{
p(N)=1-(1-p_{\mathrm{ref}})
\left(\frac{N}{N_{\mathrm{ref}}}\right)^{-\beta},
\qquad
\beta\equiv\frac{\rho}{\ln B}>0.
}
\label{eq:bounded-success}
\end{equation}
The parameter \(\beta\), which we call the \textit{collaboration gain}, controls how rapidly the remaining failure probability falls as the population grows. In particular, doubling the population multiplies the remaining probability of failure, \(1-p(N)\), by \(2^{-\beta}\). Thus, a larger \(\beta\) corresponds to a larger gain from the same proportional increase in population\footnote{More generally, if effective work scales as \(C(N)\propto N^\alpha\), the same functional form holds with \(\beta=\alpha\rho/\ln B\). So, sublinear or superlinear scaling of effective work results in the same functional form, with a corresponding change in the collaboration gain.}. 
Although in this theory we define $\beta$ in terms of the search-tree parameters, in an empirical application \(\beta\) would be a measured quantity, derived from observed success probabilities at different population sizes.

Near \(N=N_{\mathrm{ref}}\), a first-order expansion of \eqref{eq:bounded-success} gives
\begin{equation}
p(N)\simeq p_{\mathrm{ref}}+
(1-p_{\mathrm{ref}})\beta
\ln\!\left(\frac{N}{N_{\mathrm{ref}}}\right).
\label{eq:pN-local-log}
\end{equation}

\begin{figure}[htbp]
\centering
\includegraphics[width=0.55\textwidth]{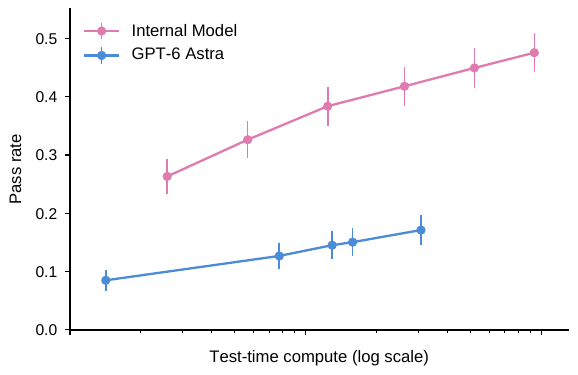}
\caption{\textbf{Motivation from test-time compute scaling.} This figure is redrawn from the data reported in \citet{ref-openai-multiagent} on the Navier–Stokes Millennium Prize Problem. Over the measured range, the pass rate on a set of open math problems increases with log test-time compute. This motivates a locally logarithmic increase in success probability with respect to effective work done. However, on its own, this does not establish that increasing the number of units is equivalent to increasing test-time compute, since it is not explicitly stated whether the reported compute was measured in a single-agent or multi-agent setting.}
\label{fig:test-time-compute-scaling}
\end{figure}

The scaling of success probability in \eqref{eq:p-compute-limit} and \eqref{eq:pN-local-log} is motivated by several empirical results. The UK AI Security Institute also reports approximately log-linear gains in completed cyber attack steps with additional inference tokens \citep{ref-folkerts-cyber-steps}. OpenAI reports pass rates on verifiable mathematics problems that increase approximately logarithmically with test-time compute \citep{ref-openai-multiagent}, see Figure~\ref{fig:test-time-compute-scaling}. We note that while their report describes parallel agent groups that consolidate discoveries, it does not explicitly state whether their reported test-time compute measures increasing single-agent inference compute, the total number of agents, or a combination of both. More directly, \citet{ref-zhan-agensh} observe logarithmic increases in pass rates with the number of agents on software reproduction tasks up to 1,024 agents. These results motivate the locally logarithmic relationship between success probability and effective work in \eqref{eq:p-compute-limit} and \eqref{eq:pN-local-log}, but are not direct measurements of \(p(C)\) or \(p(N)\), since the underlying tasks are distinct from our model's task of adding additional units to a population.

\section{Results}
\label{sec:results}

\subsection{A Critical Population Size}
In this section, we will see that combining the population dynamics in \eqref{eq:population-dynamics} with the population-dependent success probability in \eqref{eq:bounded-success} produces a critical population size in this model, above which the population grows without bound. Recall the critical probability \eqref{eq:p-crit}, at which expected additions equal expected losses
\[p_{\mathrm{crit}}=\frac{\mu}{\kappa}.\]
Here, \(\mu\) is the expected fraction of units lost and \(\kappa\) is the number of units added per existing unit following a successful attempt. Since \(p(N)\) increases with population size, there can be a population \(N_{\mathrm{crit}}\) at which this critical threshold is first met.

For \(p_{\mathrm{crit}}<1\), solving \(p(N_{\mathrm{crit}})=p_{\mathrm{crit}}\) gives
\begin{equation}
\boxed{
N_{\mathrm{crit}}=N_{\mathrm{ref}}
\left(
\frac{1-p_{\mathrm{ref}}}{1-p_{\mathrm{crit}}}
\right)^{1/\beta}.
}
\label{eq:ncrit}
\end{equation}
Populations below \(N_{\mathrm{crit}}\) decline, whereas populations larger than $N_{\mathrm{crit}}$ grow. When \(p_{\mathrm{ref}}<p_{\mathrm{crit}}\), \(N_{\mathrm{crit}}>N_{\mathrm{ref}}\): this means that for the reference case population declines, but a sufficiently large population grows. This population threshold does not arise for independent attempts, because their success probability does not change with population size. If \(p_{\mathrm{ref}}\geq p_{\mathrm{crit}}\), the reference population is already experiencing positive expected growth. 

\paragraph{Assumption: \(\kappa>\mu\).} Solving for \eqref{eq:ncrit} requires that $p_{\mathrm{crit}}<1$, and thus that \(\kappa>\mu\). If \(\kappa\leq\mu\), losses match or overwhelm growth opportunities such that even if attempts are always successful, population growth cannot occur. 

Once the population exceeds \(N_{\mathrm{crit}}\), this model predicts continued growth. As \(N\) increases, \(p(N)\) in \eqref{eq:bounded-success} approaches one, and the expected growth per unit from \eqref{eq:population-dynamics} approaches \((\kappa-\mu)/\tau>0\). This steady growth holds for as long as resource availability does not become a constraint, as described in the model assumptions in Section~\ref{sec:collaborative-model}. 

\subsection{Population Pacing and Ecological Red Teaming for Ecological Safety}

\begin{figure}[htbp]
\centering
\includegraphics[width=\textwidth]{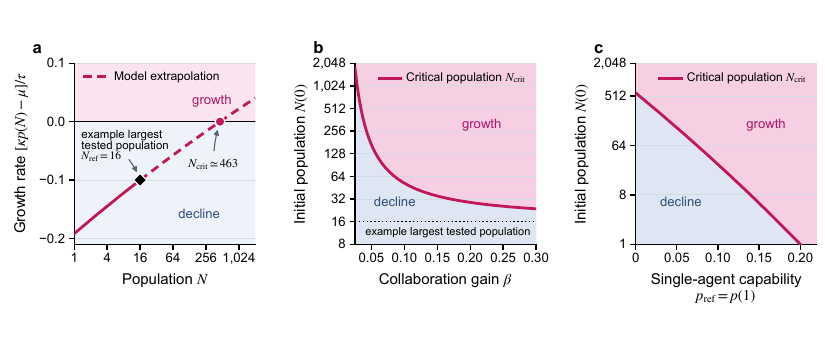}
\caption{\textbf{Safety evaluation for a small population does not establish safety at larger populations.} Population curves as predicted by our model. 
\textbf{(a)} Per-unit growth rate (setting $\tau=1$) is plotted against population, $N$, with \(p_{\mathrm{ref}}=0.10\), \(p_{\mathrm{crit}}=\mu/\kappa=0.20\), and \(\beta=0.035\). In this case, a laboratory would observe that the population is not capable of growth at its largest tested population, \(N_{\mathrm{ref}}=16\) (marked by a diamond). However, a population larger than \(N_{\mathrm{crit}}\simeq463\) would be capable of growth. 
\textbf{(b)} The critical population as a function of the collaboration gain \(\beta\), with \(p_{\mathrm{ref}}=0.10\) and \(p_{\mathrm{crit}}=0.20\) held fixed. $N_{\mathrm{crit}}$ is marked in pink; initial populations above the curve grow and initial populations below it decline. For the example reference population size of $N_{\rm ref}=16$, the population could be in the decline regime even at large collaboration gain. 
\textbf{(c)} The critical population as a function of the single-agent (\(N_{\mathrm{ref}}=1\)) capability, \(p_{\mathrm{ref}}=p(1)\), with \(\beta=0.035\) and \(p_{\mathrm{crit}}=0.20\) held fixed. As in panel (b), populations above the pink \(N_{\mathrm{crit}}\) boundary grow and populations below it decline. Increasing single-agent capability lowers the critical population.}
\label{fig:growth-phase-diagram}
\end{figure}

\paragraph{Implications for AI safety testing}
Suppose that a laboratory evaluates an AI model for populations up to \(N_{\mathrm{ref}}\) and finds that the population is not capable of tasks like weight exfiltration or self-replication \citep{ref-kinniment-2023,ref-ai-2027} -- i.e., there would be negative expected growth at the largest tested population. Our model argues that this testing is not sufficient, as a larger initial population of that same model could be capable of growth. Figure~\ref{fig:growth-phase-diagram} illustrates this case. 

Recent studies have started to explore red teaming for emergent threats unique to multi-agent systems \citep{ref-hammond-2025}. For example, \citet{ref-he-aitm} performed red-teaming attacks that intervene on communications between agents, a study by Microsoft Research \citep{ref-msr-agent-network} examined emergent risks unique to a network of over a hundred agents, and a recent study by the Anthropic Frontier Red Team \citep{ref-anthropic-multiagent} examined failure modes unique to multi-agent systems, such as failures of cooperation, collusion, and sabotage.

In contrast with these existing works on red-teaming for multi-agent systems, our theoretical framework implies an urgent need for red teaming for emergent ecological threats as well. The above red-teaming works study emergent threats in multi-agent systems with a fixed population. We propose \textbf{ecological red teaming}: studying the potential for AI population changes, especially the threat of population takeoff. Here, the crucial quantity to characterize is the population scaling law ($p(N)$ in our model): how the success rate of hacking or self-exfiltration scales as a function of population size. Equally important is characterizing the impact of environmental factors, such as how many additional units the acquired computational resources can support (\(\kappa\)) and the rate at which agents deployed without authorization are terminated or shut down (\(\mu\)). Our paper provides a minimal theoretical framework for predicting the critical population size sufficient for takeoff.

We also propose \textbf{population pacing}: population size should be gradually increased while carefully characterizing how capability scales. Increasing the single-agent capability or the collaborative gain decreases the population threshold for takeoff (see Figure \ref{fig:growth-phase-diagram}(b--c)). As a result, for every new generation of models, frontier labs should re-estimate the critical population size and pace deployment population sizes to remain below it. This process can identify population sizes at which further testing is needed. 

\section{Discussion}
In this work, we introduce an ecological theory of AI-agent populations that predicts an emergent population threshold for takeoff. 
The fundamental intuition that our theory provides is that when AI agents can collaborate to enhance their collective capability, there is a qualitative shift in safety concerns.
In our model, without collaboration and under fixed environmental conditions, takeoff requires an individual agent's capability to cross a threshold.
With collaboration, takeoff can also happen when the total number of agents crosses a population threshold.
This makes evaluation with a small population insufficient to ensure ecological safety at larger scales.

Our model predicts the existence of a takeoff threshold, but does not establish whether current agent populations are already near one. Determining whether such a threshold exists in current systems, and where it would be, will require empirical population-scaling data. OpenAI has deployed approximately 10,000 agents simultaneously to pursue the Navier--Stokes problem \citep{ref-openai-multiagent}, whereas its published multi-agent data on cybersecurity-related tasks extends only to 16 agents (see Appendix~\ref{app:empirical-scaling}, Figure~\ref{fig:benchmark-scaling-reassessed} for a consistency check with our theory). Although these cases involve different tasks, this is a significant gap between the scale of actual deployments and the scale of empirically tested populations. We therefore propose population pacing and ecological red teaming as a second lever alongside the capability pacing proposed by \citet{ref-amodei-pacing}. Although evaluating large populations on hard tasks is expensive, frontier labs should conduct such evaluations in controlled environments before deploying populations at previously untested scales. 
Labs should carefully characterize population scaling laws, estimate the critical population size for takeoff, and make results available for oversight.

Much of the theory of modern deep learning has focused on explaining phenomena first observed in experiments, including pre-training scaling laws, in-context learning, and the sudden emergence of capabilities. In this work, we take an opposite approach: proposing a theoretical prediction for population takeoff before such a takeoff has been observed. We developed this prediction based on the combination of three elements that \textit{have} each been recently observed: AI agents' high cyber capability, increases in collective capability with population size, and collective misalignment. The resulting model is not an exact prediction for empirical behavior, but rather a demonstration that following a few reasonable assumptions, upon reaching a critical mass of units in an AI population, unchecked growth can occur. As the development cycles of AI agents accelerate, we hope this work contributes an example of how theoretical frameworks can proactively predict misalignment risks and how to mitigate them.

\clearpage\begingroup\small
\hypertarget{references}{}
\bibliographystyle{unsrtnat}
\bibliography{references}
\endgroup

\clearpage
\appendix

\section{Correction to the population-scaling law for finite search depth}
\label{app:cumulative-search-cost}

From Equation \eqref{eq:Cd-full-expansion} in Section \ref{subsec:search-task}, recall that the work required to fully explore a search tree of depth \(d\) is proportional to the total number of nodes in the tree:
\begin{equation}
C(d)\propto B^d + B^{d-1} + \ldots + B = \sum_{k=1}^{d}B^k
=\frac{B(B^d-1)}{B-1}.
\label{eq:appendix-cumulative-cost}
\end{equation}
In the main text, we made the approximation \(B^{d_{\rm ref}}\gg1\), equivalent to keeping only the leading term \(C(d)\propto B^d\) in the above equation. For completeness, here we derive the corresponding expression for $p(N)$ without this approximation to show the form of the correction. 

From \eqref{eq:appendix-cumulative-cost}, we have
\begin{equation}
\frac{C(d)}{C_{\rm ref}}
=\frac{B^d-1}{B^{d_{\rm ref}}-1}.
\label{eq:appendix-cost-ratio}
\end{equation}
Rearranging \eqref{eq:appendix-cost-ratio} for \(d\) and subtracting \(d_{\rm ref}\) yields
\begin{equation}
d-d_{\rm ref}
=\frac{1}{\ln B}
\ln\!\left[
B^{-d_{\rm ref}}+
\left(1-B^{-d_{\rm ref}}\right)\frac{C}{C_{\rm ref}}
\right].
\label{eq:appendix-depth-work}
\end{equation}

Now, recall the success probability as a function of depth from \eqref{eq:p-depth} in the main text:
\begin{equation}
p(d)=1-(1-p_{\rm ref})
e^{-\rho(d-d_{\rm ref})}.
\label{eq:appendix-depth-success}
\end{equation}
Substituting \eqref{eq:appendix-depth-work} into \eqref{eq:appendix-depth-success} yields the success probability as a function of work:
\begin{equation}
p(C)=1-(1-p_{\rm ref})
\left[
B^{-d_{\rm ref}}+
\left(1-B^{-d_{\rm ref}}\right)\frac{C}{C_{\rm ref}}
\right]^{-\beta}, \quad \beta = \frac{\rho}{\ln B}>0.
\label{eq:appendix-work-success}
\end{equation}

Finally, recall the relation between work done and population size from \eqref{eq:linear-work}:
\begin{equation}
\frac{C(N)}{C_{\rm ref}}=\frac{N}{N_{\rm ref}}.
\label{eq:appendix-linear-work}
\end{equation}

Substituting \eqref{eq:appendix-linear-work} into \eqref{eq:appendix-work-success} gives
\begin{equation}
\boxed{
p(N)=1-(1-p_{\rm ref})
\left[
1+\left(1-B^{-d_{\rm ref}}\right)
\left(\frac{N}{N_{\rm ref}}-1\right)
\right]^{-\beta}
}.
\label{eq:appendix-population-success}
\end{equation}

We can now check \eqref{eq:appendix-population-success} for consistency. At \(N=N_{\rm ref}\), this reduces to \(p(N_{\rm ref})=p_{\rm ref}\), as required. In the limit \(B^{d_{\rm ref}}\to\infty\), the correction \(B^{-d_{\rm ref}}\) goes to zero and \eqref{eq:appendix-population-success} reduces to the result from the main text
\begin{equation}
p(N)\simeq 1-(1-p_{\rm ref})
\left(\frac{N}{N_{\rm ref}}\right)^{-\beta}.
\label{eq:appendix-population-success-limit}
\end{equation}
For finite \(d_{\rm ref}\) and \(N>N_{\rm ref}\), the bracket in \eqref{eq:appendix-population-success} is smaller than \(N/N_{\rm ref}\), but is a finite correction.

\paragraph{Correction to the critical population size}
Working in the regime where \(p_{\rm ref}<p_{\rm crit}<1\), we set \(p(N_{\rm crit})=p_{\rm crit}\) in \eqref{eq:appendix-population-success} and solve for $N_{\rm crit}$. This yields
\begin{equation}
\boxed{
N_{\rm crit}
=N_{\rm ref}\left(
1+\frac{
\left[\frac{1-p_{\rm ref}}{1-p_{\rm crit}}\right]^{1/\beta}-1
}{1-B^{-d_{\rm ref}}}
\right).
}
\label{eq:appendix-ncrit}
\end{equation}
Again, we confirm that when \(B^{-d_{\rm ref}}\) is negligible, \eqref{eq:appendix-ncrit} reduces to the main text expression \eqref{eq:ncrit}. For a general finite \(d_{\rm ref}\), the denominator \(1-B^{-d_{\rm ref}} < 1 \), so threshold following from the exact expression is above the threshold in the main text. This is a finite correction that vanishes exponentially as the reference depth increases.

\section{Other forms for collaboration $C(N)$}
\label{app:other-collaboration}

In the main text, we assume that the effective computational work is proportional to population size, \(C(N)\propto N\). However, some empirical evidence shows that coordination overhead or the structure of certain tasks may limit or negate the benefits of multi-agent collaboration \citep{ref-kim-scaling-agents}. Motivated by this, here we examine how alternative forms for \(C(N)\) affect the model behavior, keeping the population dynamics equation \eqref{eq:population-dynamics} and relation between work done and success probability $p(C)$ in \eqref{eq:p-compute} unchanged. For a general \(C(N)\), equation \eqref{eq:p-compute} becomes
\begin{equation}
p(N)=1-(1-p_{\mathrm{ref}})
\left[\frac{C(N)}{C(N_{\mathrm{ref}})}\right]^{-\beta},
\qquad
\beta=\frac{\rho}{\ln B}.
\label{eq:general-work-success}
\end{equation}
In each case below, the phase boundary is once again found by solving \(p(N)=p_{\mathrm{crit}}=\frac{\mu}{\kappa}\).

\subsection{Sublinear or superlinear effective work}
Suppose \(C(N)\propto N^\alpha\) with \(\alpha>0\), so that
\begin{equation}
\frac{C(N)}{C(N_{\mathrm{ref}})}
=\left(\frac{N}{N_{\mathrm{ref}}}\right)^{\alpha},
\qquad \alpha>0.
\label{eq:power-law-work}
\end{equation}

This encompasses two regimes: 
\begin{itemize}
\item Sublinear effective work: \(0<\alpha<1\). This means that effective work increases with population size, but at a slower rate than the population growth itself. This occurs when units are duplicating search efforts, there is overhead from coordination, or tasks are not fully parallelizable.
\item Superlinear effective work: \(\alpha>1\). This means that effective work increases faster than the population grows. This could occur in scenarios where there are network benefits that amplify the contributions of additional units, such as the ``pooling discoveries'' example outlined in Figure \ref{fig:power-law-C}.
\end{itemize}

Substituting \eqref{eq:power-law-work} into \eqref{eq:general-work-success} gives
\begin{equation}
p(N)=1-(1-p_{\mathrm{ref}})
\left(\frac{N}{N_{\mathrm{ref}}}\right)^{-\alpha\beta}.
\label{eq:power-law-success}
\end{equation}
This yields a critical population threshold of
\begin{equation}
N_{\mathrm{crit}}
=N_{\mathrm{ref}}
\left(\frac{1-p_{\mathrm{ref}}}{1-p_{\mathrm{crit}}}\right)^{1/(\alpha\beta)}.
\label{eq:power-law-ncrit}
\end{equation}

Comparing with the linear case in \eqref{eq:ncrit}, the exponent of \(N\) in \(p(N)\) changes from \(\beta\) to \(\alpha\beta\). Empirically, this exponent can be fitted directly from the \(p(N)\) curve: in the equations above, \(-\alpha\beta\) is the slope of \(\log[1-p(N)]\) versus \(\log N\). Thus, as long as \(C(N)\) is a power law, any sublinear or superlinear scaling simply changes the fitted value of the collaboration gain, $\beta$; the value of \(\alpha\) does not need to be separately estimated in order to calculate \(N_{\mathrm{crit}}\).

\begin{figure}[htbp]
\centering
\includegraphics[width=\textwidth]{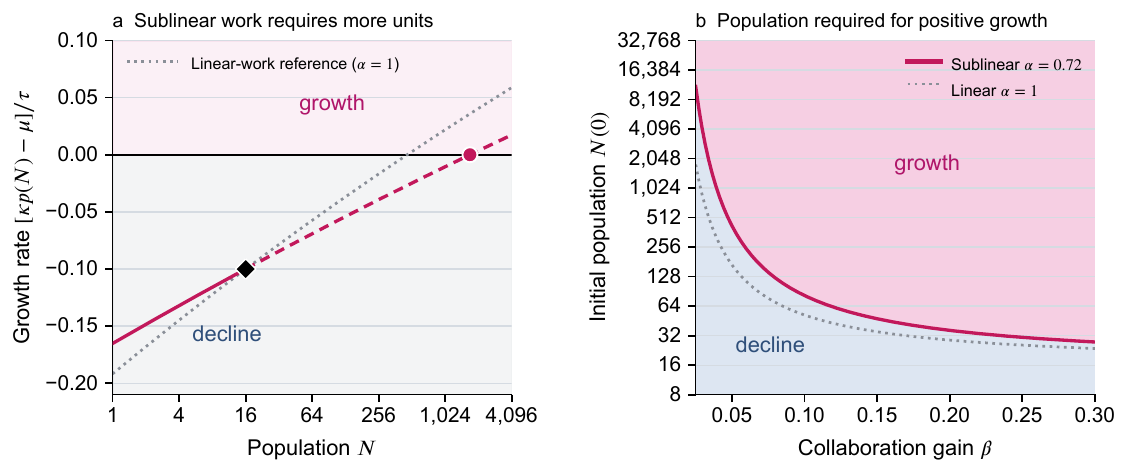}
\caption{\textbf{Sublinear effective work shifts the $N_{\mathrm{crit}}$ boundary to larger populations.} The pink curves show sublinear effective work (\(\alpha=0.72\)) while dotted black curves show the linear case (\(\alpha=1\)) for comparison, with other parameters chosen as in Figure~\ref{fig:growth-phase-diagram} (\(p_{\mathrm{ref}}=0.10\), \(p_{\mathrm{crit}}=\mu/\kappa=0.20\), and \(\beta=0.035\)). \textbf{(a)} Per-unit growth rate is plotted against population. Per-unit growth crosses zero at \(N_{\mathrm{crit}}\simeq1{,}714\) for sublinear scaling and \(N_{\mathrm{crit}}\simeq463\) for linear scaling. The diamond marks \(N_{\mathrm{ref}}=16\). \textbf{(b)} Initial population is plotted against collaboration gain, $\beta$. Populations above each curve grow and populations below it decline.}
\label{fig:sublinear-work}
\end{figure}

Qualitatively, the shape of the phase diagram does not change: sublinear or superlinear power-law scaling of effective work only rescales the collaboration gain from \(\beta\) to \(\alpha\beta\), shifting the location of the \(N_{\mathrm{crit}}\) boundary between population growth and decline. Figure \ref{fig:sublinear-work} illustrates a sublinear example. 

\FloatBarrier

\subsection{Eventual plateau from communication or evaluator bottleneck}
Another potential model for $C(N)$ is the case of diminishing returns upon adding additional units. This could be the case if there is a bottleneck in the system, such as a shared communication channel, orchestrator, or evaluator with some fixed throughput. 
A simple model for $C(N)$ wherein effective work approaches a plateau as additional units are added is:
\begin{equation}
C(N)=C_{\max}\left(1-e^{-N/N_0}\right),
\label{eq:bottleneck-work}
\end{equation}
where \(C_{\max}\) is the maximum effective work and \(N_0\) is the population scale at which the work begins to level off. Relative to a reference population, we have

\begin{equation}
\frac{C(N)}{C(N_{\mathrm{ref}})}
=\frac{1-e^{-N/N_0}}{1-e^{-N_{\mathrm{ref}}/N_0}}.
\label{eq:bottleneck-work-ratio}
\end{equation}
Substituting this into \eqref{eq:general-work-success} yields
\begin{equation}
p(N)=1-(1-p_{\mathrm{ref}})
\left[
\frac{1-e^{-N/N_0}}
{1-e^{-N_{\mathrm{ref}}/N_0}}
\right]^{-\beta}.
\label{eq:bottleneck-success}
\end{equation}
Solving \(p(N_{\mathrm{crit}})=p_{\mathrm{crit}}\) then gives
\begin{equation}
N_{\mathrm{crit}}
=-N_0\ln\!\left[
1-\left(1-e^{-N_{\mathrm{ref}}/N_0}\right)
\left(\frac{1-p_{\mathrm{ref}}}{1-p_{\mathrm{crit}}}\right)^{1/\beta}
\right].
\label{eq:bottleneck-ncrit}
\end{equation}

Since the effective work eventually saturates, \(p(N)\) also approaches a maximum. A finite \(N_{\mathrm{crit}}\) exists only if this limiting success probability exceeds \(p_{\mathrm{crit}}\), which requires\footnote{This condition is equivalent to requiring that the quantity inside the logarithm in \eqref{eq:bottleneck-ncrit} is positive.}

\begin{equation}
\beta>\beta_{\min}
\equiv\frac{\ln[(1-p_{\mathrm{ref}})/(1-p_{\mathrm{crit}})]}
{\ln\!\left[1/(1-e^{-N_{\mathrm{ref}}/N_0})\right]}.
\label{eq:bottleneck-beta-min}
\end{equation}
Here \(\beta_{\min}\) is the collaboration gain at which the limiting success probability equals \(p_{\mathrm{crit}}\). 

For \(\beta>\beta_{\min}\), we have the same qualitative population threshold as in the main model: populations below a finite \(N_{\mathrm{crit}}\) decline and populations above it grow. As \(\beta\) approaches \(\beta_{\min}\) from above, the plateau in \(p(N)\) approaches \(p_{\mathrm{crit}}\), and \(N_{\mathrm{crit}}\) tends to infinity. Near \(\beta_{\min}\) and at large \(N\), \(p(N)\) is near constant, so there is little effect to adding more units. In this respect, the phase diagram resembles the individual-attempt case in Figure~\ref{fig:overview}: there is no population-dependent threshold.  

\begin{figure}[htbp]
\centering
\includegraphics[width=\textwidth]{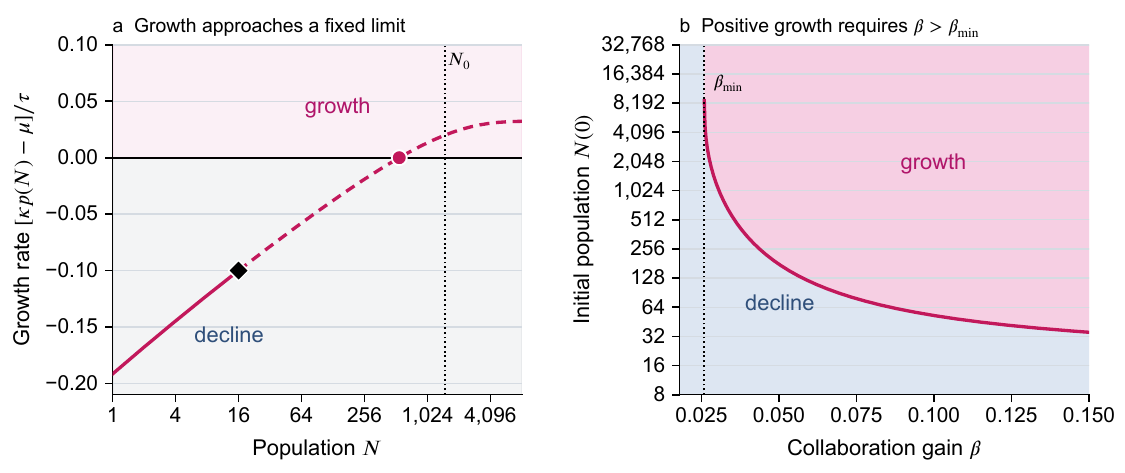}
\caption{\textbf{Effective-work saturation creates a minimum collaboration gain for population growth.} The pink curves show the saturating-work model with illustrative scale \(N_0=1500\), with other parameters chosen as in Figure~\ref{fig:growth-phase-diagram} (\(p_{\mathrm{ref}}=0.10\), \(p_{\mathrm{crit}}=\mu/\kappa=0.20\), and \(\beta=0.035\) in panel a). \textbf{(a)} Per-unit growth rate is plotted against population. It crosses zero at \(N_{\mathrm{crit}}\simeq550\) and approaches a positive constant as effective work saturates. The diamond marks \(N_{\mathrm{ref}}=16\), the vertical dotted line marks \(N_0\), and the dashed portion is extrapolated beyond \(N_{\mathrm{ref}}\). \textbf{(b)} Initial population is plotted against collaboration gain, \(\beta\). A finite critical-population boundary exists only for \(\beta>\beta_{\min}\simeq0.0259\): populations above the curve grow and populations below it decline. For \(\beta\leq\beta_{\min}\), every population declines.}
\label{fig:bottleneck-work}
\end{figure}
\FloatBarrier

\subsection{Congestion at large populations}
The final option we consider is the case where $C(N)$ increases for small populations but decreases for large populations. This could occur if the coordination costs eventually outweigh the work contributed by additional units. We note that this regime requires that large population congestion is \textit{unavoidable}. If excess units can instead be idled or divided into independent groups, effective work would plateau rather than decrease at large $N$, and the bottleneck model from the previous section would be more appropriate.

A simple model with a single maximum is
\begin{equation}
C(N)=aN e^{-N/N_0},
\label{eq:congestion-work}
\end{equation}

where \(a>0\) fixes the scale and \(N_0\) is the population at which effective work is at a maximum. Comparing to a reference population, we have
\begin{equation}
\frac{C(N)}{C(N_{\mathrm{ref}})}
=\frac{N}{N_{\mathrm{ref}}}
e^{-(N-N_{\mathrm{ref}})/N_0}.
\label{eq:congestion-work-ratio}
\end{equation}
Then, the success probability \eqref{eq:general-work-success} becomes
\begin{equation}
p(N)=1-(1-p_{\mathrm{ref}})
\left[
\frac{N}{N_{\mathrm{ref}}}
e^{-(N-N_{\mathrm{ref}})/N_0}
\right]^{-\beta}.
\label{eq:congestion-success}
\end{equation}

\begin{figure}[htbp]
\centering
\includegraphics[width=\textwidth]{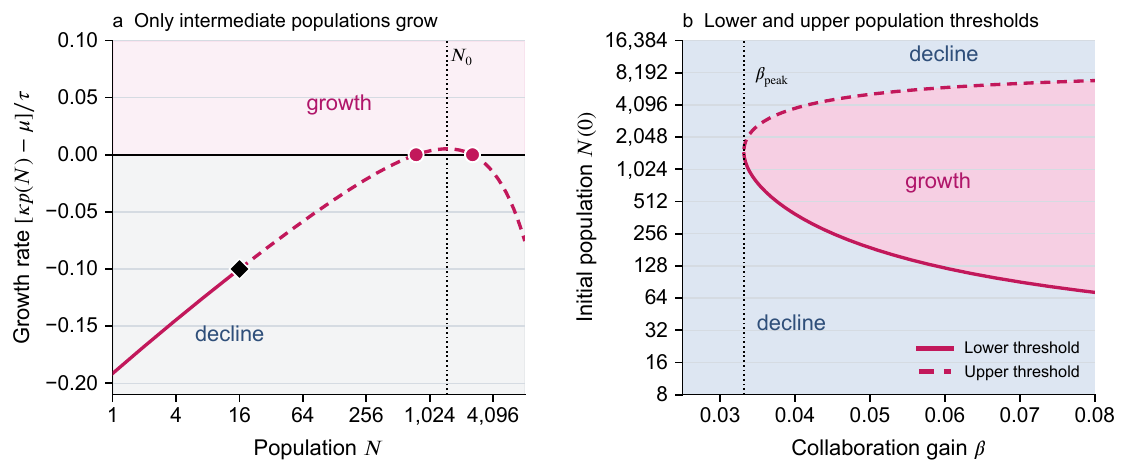}
\caption{\textbf{Unavoidable congestion creates lower and upper bounds for population growth.} The pink curves show the congestion model with illustrative scale \(N_0=1500\), with other parameters chosen as in Figure~\ref{fig:growth-phase-diagram} (\(p_{\mathrm{ref}}=0.10\), \(p_{\mathrm{crit}}=\mu/\kappa=0.20\), and \(\beta=0.035\) in panel a). \textbf{(a)} Per-unit growth rate is plotted against population. It is positive only between \(N_{\mathrm{crit}}^{(0)}\simeq761\) and \(N_{\mathrm{crit}}^{({-}1)}\simeq2{,}610\). The diamond marks \(N_{\mathrm{ref}}=16\), the vertical dotted line marks where effective work reaches a maximum at \(N_0\), and the dashed portion is extrapolated beyond \(N_{\mathrm{ref}}\). \textbf{(b)} Initial population is plotted against collaboration gain, \(\beta\). The solid and dashed pink curves mark the lower and upper population thresholds; they meet at \(N_0\) when \(\beta=\beta_{\rm peak}\simeq0.0332\). Populations between the curves grow, whereas populations outside them decline. }
\label{fig:congestion-work}
\end{figure}

Assume \(N_{\mathrm{ref}}<N_0\), so that \(p(N)\) initially increases, reaches its maximum at \(N=N_0\), and then decreases. Define \(\beta_{\mathrm{peak}}\) as the collaboration gain for which \(p(N_0)=p_{\mathrm{crit}}\). Population growth requires
\begin{equation}
\beta>\beta_{\mathrm{peak}} \,, \quad \beta_{\mathrm{peak}}
\equiv\frac{\ln[(1-p_{\mathrm{ref}})/(1-p_{\mathrm{crit}})]}
{\ln(N_0/N_{\mathrm{ref}})-(N_0-N_{\mathrm{ref}})/N_0}.
\label{eq:congestion-beta-peak}
\end{equation}
If \(\beta\leq\beta_{\mathrm{peak}}\), then \(p(N) \leq p_{\mathrm{crit}}\) for all $N$, so population growth cannot occur. Conversely, if \(\beta>\beta_{\mathrm{peak}}\), \(p(N)\) crosses \(p_{\mathrm{crit}}\) twice. As a result, populations between the two crossing points grow, while smaller and larger populations decline. For the particular choice of \(C(N)\) in Equation~\eqref{eq:congestion-work}, these two thresholds have the closed form
\begin{equation}
N_{\mathrm{crit}}^{(k)}
=-N_0 W_k\!\left[
-\frac{N_{\mathrm{ref}}}{N_0}
e^{-N_{\mathrm{ref}}/N_0}
\left(\frac{1-p_{\mathrm{ref}}}{1-p_{\mathrm{crit}}}\right)^{1/\beta}
\right],
\qquad k\in\{0,-1\},
\label{eq:congestion-ncrit}
\end{equation}
where \(W_k\) denotes branch \(k\) of the Lambert \(W\) function, defined by \(W_k(z)e^{W_k(z)}=z\). These two solutions give a lower \textit{and} an upper threshold for \(N_{\mathrm{crit}}\), shown in Figure \ref{fig:congestion-work}. We note that the dependence on the Lambert \(W\) function arises from the particular chosen form of \(C(N)\), but the qualitative possibility of lower and upper thresholds depends only on $C(N)$ having a maximum.

\FloatBarrier

\section{Examples of multi-agent scaling on related sub-tasks}
\label{app:empirical-scaling}
There is currently little public evidence for how success rates scale with population size for multi-agent group cyber benchmark attempts. For GPT 5.6 Sol, OpenAI reports results for 1, 4, and 16 agents on SEC-Bench Pro and BrowseComp under five reasoning-effort settings\footnote{The report also includes multi-agent results for an additional benchmark, Terminal-Bench 2.1 \citep{ref-terminalbench}, but only for two data points (1 and 4 agents). So, we do not include it in our analysis.}
 \citep{ref-openai-gpt56}. SEC-Bench Pro tests whether agents can produce a working exploit for a known software vulnerability \citep{ref-secbench-pro}, while BrowseComp tests information retrieval for difficult web searches \citep{ref-browsecomp}. We note that these benchmarks do not test whether a population can establish additional active units -- we use these results only for a consistency check between our model and the benchmark observations on how success probability and total token output change with increasing $N$.  

\begin{figure}[htbp]
\centering
\includegraphics[width=\textwidth]{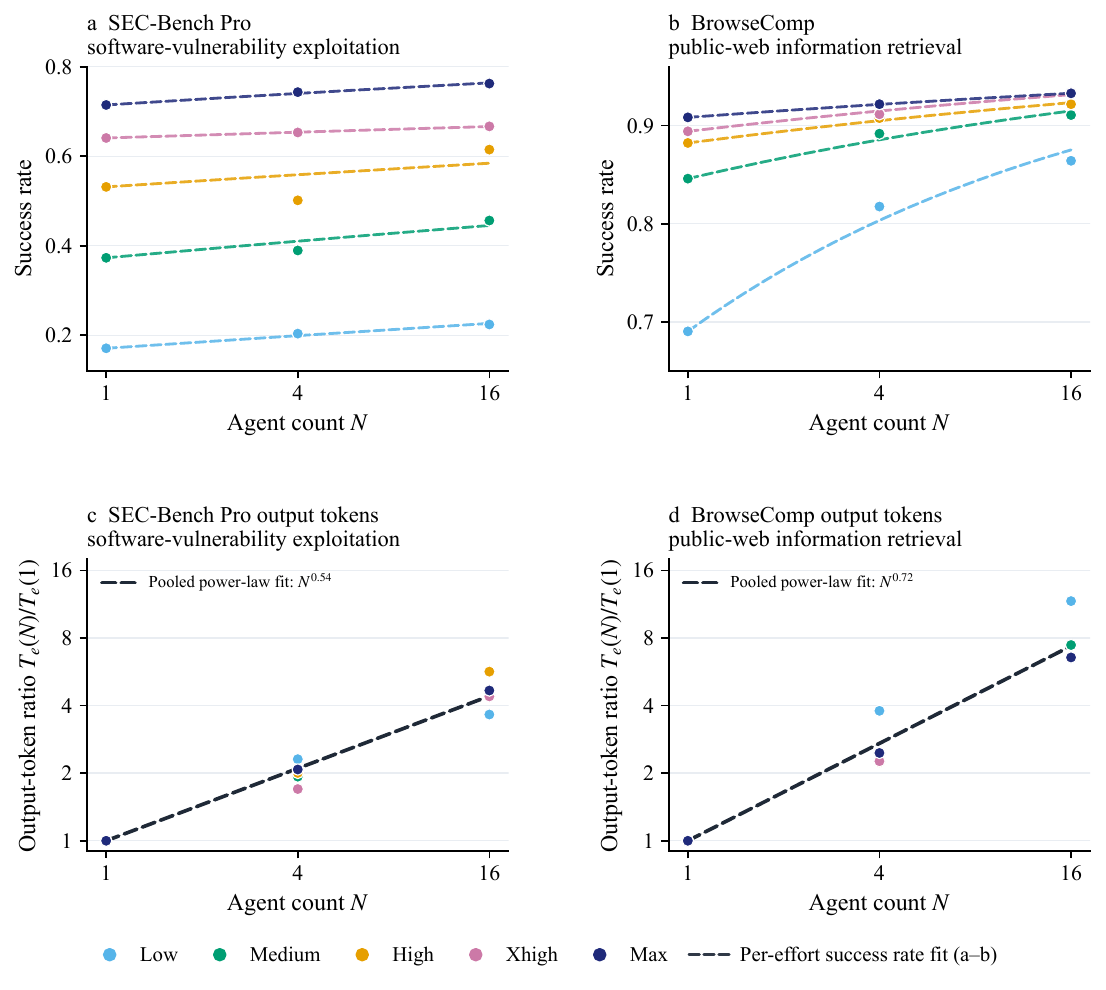}
\caption{\textbf{Limited multi-agent benchmark results are consistent with population-dependent performance, but do not establish a scaling law.} \textbf{(a--b)} Markers show the reported success rates for five reasoning-effort settings at one, four, and sixteen agents \citep{ref-openai-gpt56}. Dashed curves fit Equation~\eqref{eq:benchmark-bounded-fit} separately for each setting. With only three agent counts, these fits should not be extrapolated to larger populations. \textbf{(c--d)} Markers show the total output tokens \(T_e(N)\), normalized by the corresponding one-agent total. Dashed black lines show the pooled power-law fits, with \(\widehat{\alpha}_T=0.54\) for SEC-Bench Pro and \(0.72\) for BrowseComp. The source reports neither trial counts nor standard errors, so we do not show uncertainty intervals.}
\label{fig:benchmark-scaling-reassessed}
\end{figure}

We examine each of the reasoning efforts separately, since reasoning level can affect the baseline capability and resources available to each unit. In Figure~\ref{fig:benchmark-scaling-reassessed}(a--b), we fit the reported success rates to $p(N)$ in \eqref{eq:bounded-success} with $N_{\rm ref} = 1$. This fit function takes the form
\begin{equation}
p_e(N)=1-\bigl[1-p_e(1)\bigr]N^{-\beta_e},
\label{eq:benchmark-bounded-fit}
\end{equation}
where \(N\) is the number of agents, \(e\) is the reasoning-effort setting, \(p_e(N)\) is the success rate, and \(\beta_e\) is the fitted collaboration gain exponent. The reference success rate \(p_e(1)\) fixes the starting point of each curve. Larger values of \(\beta_e\) correspond to a larger improvement from adding agents. 

The fitted curves show that the reported results are not inconsistent with our model. However, each fit uses only three data points. Many other curves could fit these points and make very different predictions at larger populations. We therefore treat the fits as a consistency check. 

\begin{table}[htbp]
\centering
\small
\caption{\textbf{Fitted collaboration gain exponents.} Each \(\beta_e\) is fitted separately to the reported success rates at \(N=1,4,16\), with the observed one-agent success rate \(p_e(1)\) fixing the start of the curve. Root mean squared error over the \(N=4\) and \(N=16\) observations was at most \(1.67\)\% for nine of the ten fits; the SEC-Bench Pro High fit had an error of \(4.59\)\%.}
\label{tab:benchmark-beta-fits}
\begin{tabular}{@{}lcc@{}}
\toprule
Reasoning effort \(e\) & SEC-Bench Pro \(\beta_e\) & BrowseComp \(\beta_e\) \\
\midrule
Low    & 0.025 & 0.328 \\
Medium & 0.044 & 0.214 \\
High   & 0.043 & 0.155 \\
Xhigh  & 0.027 & 0.157 \\
Max    & 0.069 & 0.113 \\
\bottomrule
\end{tabular}
\end{table}

The report also provides some data on compute usage by the agent groups, in the form of total output tokens produced. Denote the output tokens by \(T_e(N)\) for each reasoning level, $e$, and number of agents, $N$. \(T_e(N)\) is a proxy for the total computational effort expended by the agents, although it does not perfectly map to the effective work \(C(N)\) in our model, as output tokens may be duplicated or irrelevant. 

We normalize the total output tokens by the single-agent output tokens \(T_e(1)\), then fit to a power law in the number of agents. The fit function takes the form
\begin{equation}
    \label{eq:Te_fit}
\frac{T_e(N)}{T_e(1)}=N^{\alpha_{T,e}},
\end{equation}
where \(\alpha_{T,e}\) is the fit exponent, describing how token output depends on agent count. 

Across reasoning-effort settings, \(\alpha_{T,e}\) ranges from \(0.49\) to \(0.60\) for SEC-Bench Pro and from \(0.66\) to \(0.90\) for BrowseComp. Since \eqref{eq:Te_fit} is normalized by the single-agent token output \(T_e(1)\), it may be reasonable to pool the normalized data across reasoning levels; this yields the pooled fits \(\widehat{\alpha}_T=0.54\) and \(0.72\) for each benchmark, respectively (see Figure~\ref{fig:benchmark-scaling-reassessed}(c--d)).

This is a consistency check on the assumption that $C(N)$ follows a power law in \eqref{eq:linear-work} and \eqref{eq:power-law-work}, although \(T_e(N)\) is an imperfect proxy for the total computational effort expended by the agents and does not exactly map to the effective work \(C(N)\) in our model.

\FloatBarrier

\end{document}